\documentclass[11pt,letterpaper]{article}
\usepackage{naaclhlt2015}
\usepackage{times}
\usepackage{latexsym}
\usepackage{url}
\usepackage{epsfig}
\usepackage{graphicx}
\usepackage{amsmath}
\usepackage{amssymb}
\usepackage{subfigure}

\title{What's Cookin'? Interpreting Cooking Videos using Text, Speech
  and Vision}

\author{
Jonathan Malmaud, Jonathan Huang, Vivek Rathod, Nick Johnston, \\\textbf{Andrew Rabinovich, and Kevin Murphy}\\
	    Google\\
	    1600 Amphitheatre Parkway\\
	    Mountain View, CA 94043\\
      {\tt malmaud@mit.edu}\\
	    {\tt \{jonathanhuang, rathodv, nickj, amrabino, kpmurphy\}@google.com}}

\date{}

\newcommand{\eat}[1]{}

\newcommand{\todo}[1]{{\bf [ToDo: #1]}}

\begin{document}
\maketitle

\begin{abstract}
We present a novel method for aligning a sequence of instructions to a
video of someone carrying out a task.
In particular, we focus on the cooking domain, where the instructions
correspond to the recipe.
Our technique relies on an HMM
to align the recipe steps to the (automatically generated)
speech transcript. We then refine this alignment using a
state-of-the-art visual food detector, based on a deep convolutional
neural network.
We show that our technique outperforms simpler techniques based on
keyword spotting.
It also enables interesting applications, such as automatically
illustrating recipes with keyframes, and searching within a video for
events of interest.
\end{abstract}

\section{Introduction}
\label{sec:intro}

In recent years, there have been many successful attempts to build large ``knowledge bases'' (KBs), such as NELL \cite{Carlson10}, KnowItAll \cite{Etzioni11}, YAGO \cite{Suchanek08}, and
Google's Knowledge Graph/ Vault \cite{Dong14}. These KBs mostly focus on declarative facts, such as ``Barack Obama was born in Hawaii".
But human knowledge also encompasses procedural information not yet within the scope of such declarative KBs -- instructions and demonstrations of how to dance the tango, for example, or how to change a tire on your car. A KB for organizing and retrieving such procedural knowledge could be a valuable resource for helping people  (and potentially even robots -- e.g., \cite{Saxena2014,Yang2015}) learn to perform various tasks.

In contrast to declarative information, procedural knowledge tends to be inherently multimodal. In particular, both language and perceptual information are typically used to parsimoniously describe procedures, as evidenced by the large number of ``how-to'' videos and illustrated guides on the open web. To automatically construct  a multimodal database of procedural knowledge, we thus need tools for extracting information from both textual and visual sources. Crucially, we also need to figure out how these various kinds of information, which often complement and overlap each other, fit together to a form a structured knowledge base of procedures.

As a small step toward the broader goal of aligning language and perception, we focus in this paper on the problem of aligning video depictions of procedures to steps in an accompanying text that corresponds to the procedure. We focus on the cooking domain due to the prevalence of cooking videos on the web and the relative ease of interpreting their recipes as linear sequences of canonical actions. In this domain, the textual source is a user-uploaded recipe attached to the video showing the recipe's execution. The individual steps of procedures are cooking actions like
``peel an onion'', ``slice an onion'',  etc. However, our techniques can be applied to any domain that has textual instructions and corresponding videos, including videos at sites such as
\url{youtube.com}, \url{howcast.com}, \url{howdini.com} or \url{videojug.com}.

The approach we take in this paper leverages the fact that the speech signal in instructional videos is often closely related to the actions that the person is performing (which is not true in more general videos).
Thus we first align the instructional steps to the speech signal using an HMM, and then refine this alignment by using a state of the art computer vision system.

In summary, our contributions are as follows.
First, we  propose a novel system that combines text, speech and vision to perform an alignment between
textual instructions and  instructional videos.
Second,  we use our system to
create a large corpus of 180k aligned recipe-video pairs,
and an even larger corpus of 1.4M short video clips,
each labeled with a cooking action and a noun phrase. We evaluate the quality of our corpus using human raters.
Third, we show how we can use
our methods to support applications such as within-video search and recipe auto-illustration.

\section{Data and pre-processing}
\label{sec:data}

We first describe how we collected our corpus of recipes
and videos, and the pre-processing steps that we run before applying our alignment model.
The corpus of recipes, as well as the results of the alignment model,
will be made available for download at
\url{github.com/malmaud/whats_cookin}.

\subsection{Collecting a large corpus of cooking videos with recipes}
\label{sec:collectVideos}
We first searched Youtube for videos
which have been automatically tagged
with the Freebase mids
/m/01mtb (Cooking) and /m/0p57p (recipe),
and which have
(automatically produced)  English-language  speech transcripts,
which yielded a collection of 7.4M videos.
Of these videos, we kept the videos that also had accompanying descriptive text,
leaving 6.2M videos.

Sometimes the recipe for a video is included in this text description,
but sometimes it is stored on an external site.
For example, a video's text description might say
``Click here for the recipe''.  To find the recipe in such cases, we look for
sentences in the video description
with any of the following keywords:
``recipe'', ``steps'', ``cook'', ``procedure'', ``preparation'', ``method''.
If we find any such tokens, we find any URLs that are mentioned in the same sentence,
and extract the corresponding document, giving us an additional 206k documents.
We then combine the original descriptive text with any additional text that
we retrieve in this way.

Finally, in order to extract the recipe from the text description of a video,
we trained a classifier that
classifies each sentence into 1 of 3 classes:
\emph{recipe step}, \emph{recipe ingredient}, or \emph{background}.
We keep only the videos which have at least one ingredient sentence and at
least one recipe sentence. This last step leaves us with 180,000 videos.

To train the recipe classifier, we need labeled examples, which we
obtain by exploiting the fact that many text webpages
containing recipes use the
machine-readable markup defined at \url{http://schema.org/Recipe}.
From this we extract 500k examples of recipe sentences,
and 500k examples of ingredient sentences.
We also sample 500k sentences at random from webpages to represent the
non-recipe class.
Finally, we train a 3-class na\"{i}ve Bayes model on this data using simple bag-of-words feature
vectors.  The performance of this model on a separate test set is shown in
Table~\ref{tab:recipeClassifier}.

\begin{table}
\begin{center}
{\footnotesize
\begin{tabular}{llll}
Class & Precision & Recall & F1 \\ \hline
Background & 0.97 & 0.95 & 0.96 \\
Ingredient & 0.93 & 0.95 & 0.94 \\
Recipe step & 0.94 & 0.95 & 0.94
\end{tabular}\vspace{-3mm}
}
\end{center}
\caption{\footnotesize Test set performance of  text-based recipe classifier.}
\label{tab:recipeClassifier}
\end{table}

\subsection{Parsing the recipe text}
\label{sec:parseRecipe}

\eat{
\begin{figure}[!t]
\centering
\includegraphics[width=0.49\textwidth]{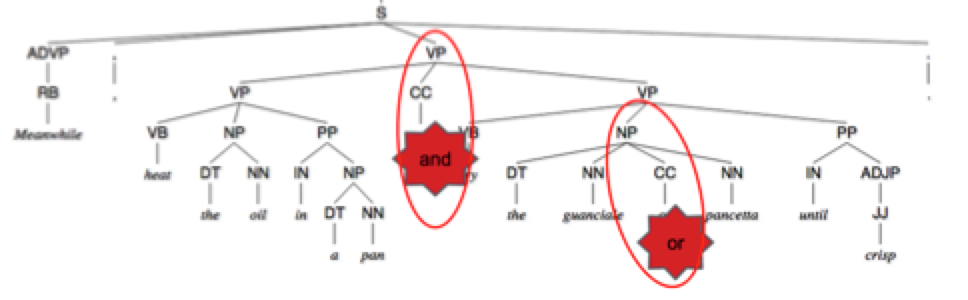}
\caption{Parse tree for the sentence "Meanwhile, heat the oil in a pan and fry the guianciale or pancetta until crisp", from
\url{https://www.youtube.com/watch?v=3AAdKl1UYZs}.
}
\label{fig:parseTree}
\end{figure}
}

\eat{
Each document is processed by a standard NLP pipeline.
This includes a part-of-speech (POS) tagger and dependency parser, comparable in accuracy to the current Stanford dependency parser \cite{klein2003accurate}, and an NP extractor that makes use of POS tags and dependency edges to identify a set of entity mentions.
Thus we separate the type tagging task from the identification of entity mentions, often performed jointly by entity recognition systems.
We also run a coreference resolver comparable to that of Haghighi and Klein \shortcite{haghighi2009simple} that clusters mentions into entities.
Lastly, our entity resolver links entity mentions to Freebase
profiles; the system maps string aliases (``Barack Obama'', ``Obama'',
``Barack H. Obama'', etc.) to profiles with probabilities derived from
Wikipedia anchors.
}

For each recipe, we apply a suite
of in-house NLP tools, similar to the Stanford Core NLP pipeline.
In particular, we perform POS tagging, entity chunking,
and constituency parsing (based on a re-implementation of
\cite{Petrov06}).\footnote{
Sometimes the parser performs poorly,
because the language used in recipes is often full of imperative
sentences, such as ``Mix the flour'', whereas the parser is trained on
newswire text.
As a simple heuristic for overcoming this, we
classify any token at the beginning of a sentence as a verb if it lexically matches
a manually-defined list of cooking-related verbs.
} %
Following \cite{Druck2012}, we use the parse tree structure
to partition each sentence into ``micro steps''.
In particular,
we split at any token categorized by the parser as a conjunction only if that token's parent in the sentence's
constituency parse is a verb phrase.
Any recipe step that is missing a verb is considered noise and discarded.

We then label each recipe step with an optional action and a list of 0 or more noun chunks.
The action label is the lemmatized version of the head verb of the recipe step. We look at all chunked noun entities in the step which are the direct object of the action (either directly or via the preposition ``of'', as in ``Add a cup of flour'').

We  canonicalize these entities by computing their similarity to the list of ingredients associated with this recipe. If an ingredient is sufficiently similar, that ingredient is added to this step's entity list.
Otherwise, the stemmed entity is used.
For example, consider the step
``Mix tomato sauce and pasta''; if the recipe has a known ingredient called
``spaghetti'',  we would label the action as ``mix'' and the entities
as ``tomato sauce'' and ``spaghetti'', because of its high semantic similarity to
``pasta''. (Semantic similarity is estimated based on Euclidean
distance between word
embedding vectors computed using the method of~\cite{Mikolov2013}
trained on general web text.)

In many cases, the direct object of a transitive verb
is elided (not explicitly stated); this is known as the ``zero
anaphora'' problem.
For example, the text may say
``Add eggs and flour to the bowl. Mix well.''.
The object of the verb ``mix'' is clearly the stuff that was just
added to the bowl (namely the eggs and flour), although this is not explicitly
stated.
To handle this, we use a simple recency heuristic, and insert the
entities from the previous step to the current step.

\subsection{Processing the speech transcript}

The output of Youtube's ASR system is a sequence of time-stamped tokens,
produced by  a standard Viterbi decoding system.
We concatenate these tokens into a single long document,
and then apply our NLP pipeline to it. Note that, in
addition to errors introduced by the ASR system\footnote{
According to \cite{Liao2013},
the Youtube ASR system we used, based on using Gaussian mixture models for the
acoustic model,
 has a word error rate of about 52\% (averaged
over all English-language videos; some genres, such as news, had lower
error rates).
The newer system, which uses deep neural nets for the acoustic model,
has an average WER of 44\%; however, this was not available to us at
the time we did our experiments.
}, %
the NLP system can
introduce additional errors, because it does not work well on
text that may be ungrammatical and which is entirely devoid of
punctuation and sentence boundary markers.

To assess the impact of these combined sources of error, we also collected a much
smaller set of 480 cooking videos (with corresponding recipe text) for
which the video creator had uploaded a manually curated speech
transcript; this has no transcription errors, it contains sentence boundary
markers, and it also aligns whole phrases with the video (instead of
just single tokens).
We applied the same NLP pipeline to these manual
transcripts. In the results section, we will see that the accuracy of
our end-to-end system is indeed higher when the speech transcript is
error-free and well-formed. However, we can still get good results
using noisier, automatically produced transcripts.

\eat{
\subsection{The visual food detection system}
\label{sec:foodDetector}

As mentioned in the introduction, we use a deep CNN to identify food
in each image.
The CNN architecture is similar to the ``Inception'' model described in
\cite{Szegedy2014} which recently won the Image Net competition.
The difference is in the training data: we collected XXX images of YYY different kinds of food from the web,
and supplemented them with ZZZ images from the training set of \cite{Bossard14}.
When we evaluated on their test set we got a performance of AAA.
\todo{Andrew: fill in details}

We applied the CNN to every frame of our video collection.
We  extract the CNN scores based on the ingredients that are mentioned in the recipe.
The result is a vector of $K$ scores per frame,
$p(s_{t,1:K}|v_t)$, where $v_t$ is the $t$'th frame of the video, $s_{t,k} \in [0,1]$ is the probability
that food $k$ is
present in frame $t$,
and $K$ is the number of foods mentioned in the recipe.
An example of the score vector over time is shown in Figure~\ref{fig:foodDetectorTrace}.
(We use word embedding vectors to find ``semantically similar''
classes in cases where the mentioned ingredient does not have the same
name as one of the pretrained classes.)
}

\section{Methods}
\label{sec:methods}

In this section, we describe our system for aligning instructional
text and video.

\subsection{HMM to align recipe with ASR transcript}
\label{sec:hmm}

\begin{figure}[!t]
\centering
\includegraphics[width=0.25\textwidth]{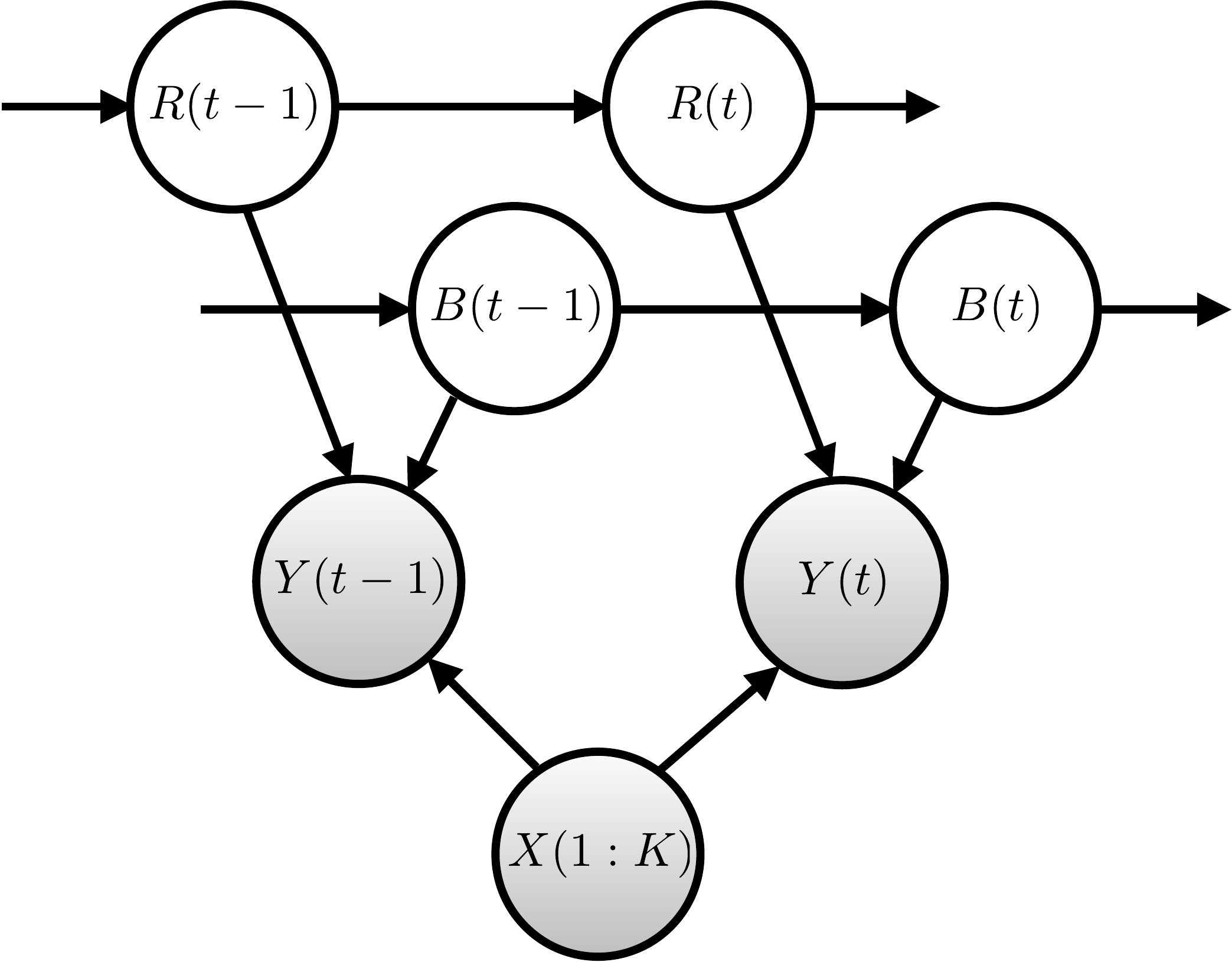}
\caption{\footnotesize Graphical model representation of the factored HMM.
See text for details.
}
\label{fig:HMM}
\end{figure}

We align each step of the recipe to a corresponding
sequence of words in the ASR transcript by using the input-output HMM
shown in Figure~\ref{fig:HMM}.
Here
$X(1:K)$ represents the textual recipe steps (obtained using the
process described in Section~\ref{sec:parseRecipe});
$Y(1:T)$ represent the ASR tokens (spoken words);
$R(t) \in \{1,\ldots,K\}$ is the recipe step number for
frame $t$;
and
 $B(t) \in \{0,1\}$ represents whether timestep $t$ is generated by
 the background ($B=1$)
or foreground model ($B=0$).  This background variable
is needed since sometimes sequences of
spoken words are unrelated to the content of the recipe, especially at
the beginning and end of a video.

The conditional probability distributions (CPDs) for the Markov chain is
as follows:\vspace{-2mm}
{\footnotesize
\begin{eqnarray*}
p(R(t)=r | R(t-1)=r')
 &=& \left\{
\begin{array}{ll}
  \alpha & \mbox{if $r=r'$+1} \\
  1-\alpha & \mbox{if $r=r'$} \\
  0.0 & \mbox{otherwise}
\end{array} \right. \\
p(B(t)= b  | B(t-1)= b) &=& \gamma.
\end{eqnarray*}
}
%
This encodes our assumption that the video follows the same ordering
as the recipe and that background/foreground tokens tend to cluster together.
Obviously these assumptions do not always hold, but they
are a reasonable approximation.

For each recipe, we set $\alpha=K/T$, the ratio of recipe steps
to transcript tokens. This setting corresponds to an \textit{a priori} belief that each
recipe step is aligned with the same number of transcript tokens. The parameter
$\gamma$ in our experiments is set by
cross-validation to $0.7$ based on a small set of manually-labeled recipes.

For the foreground observation model, we generate the
observed word from the corresponding recipe step via:
{\footnotesize
\begin{eqnarray*}
\lefteqn{\log p(Y(t)=y | R(t)=k, X(1:K), B(t)=0) \propto}  \\
& \mbox{max}(\{ \mbox{WordSimilarity}(y, x): x  \in X(k)\}),
\end{eqnarray*}
} %
where $X(k)$ is the set of words in the $k$'th recipe step,
and $\mbox{WordSimilarity}(s,t)$ is a measure of similarity between
words $s$ and $t$, based on word vector distance.

If this frame is aligned to the background, we generate it from the
empirical distribution of words, which is estimated based on pooling
all the data:\vspace{-2mm}
\begin{eqnarray*}
 p(Y(t)=y | R(t)=k, B(t)=1) = \hat{p}(y).
\end{eqnarray*}
Finally, the prior for $p(B(t))$ is uniform,
and  $p(R(1))$ is set to a delta function on $R(1)=1$
(i.e., we assume videos start at step 1 of the recipe).

Having defined the model, we ``flatten'' it to
a standard HMM (by taking the cross product of $R_t$ and $B_t$),
then estimate the MAP sequence using the Viterbi algorithm.
See Figure~\ref{fig:alignment_example} for an example.

Finally, we label each segment of the video  as
follows: use the segmentation induced by the alignment,
and extract the action and object from the corresponding recipe
step as described in Section~\ref{sec:parseRecipe}.
If the segment was labeled as background by the HMM, we do not
apply any label to it.

\subsection{Keyword spotting}
\label{sec:kw}

A simpler approach to labeling video segments is to just search for
verbs in the ASR transcript, and then to extract a
fixed-sized window around the timestamp where the keyword occurred.
We call this approach ``keyword spotting''.
A similar method from \cite{Yu14} filters ASR transcripts by part-of-speech tag
and finds tokens that match a small vocabulary
to create a corpus of video
clips (extracted from instructional videos), each labeled with an action/object pair.

In more detail, we manually define a whitelist of
$\sim$200 actions (all transitive verbs) of interest, such as ``add'', ``chop'',
``fry'', etc. We then identify when these words are spoken (relying
on the POS tags to filter out non-verbs), and
extract an 8 second video clip around this timestamp.
(Using 2 seconds prior to the action being mentioned, and 6 seconds
following.)
To extract the object, we take all tokens tagged as ``noun''
within 5 tokens after the action.

\subsection{Hybrid HMM + keyword spotting}
\label{sec:hybrid}

\eat{
To assemble our clip database, we used a procedure that attempts combined the strengths
of the HMM model and keyword spotting. In particular, keyword spotting results in approximately 5
times as many clips as relying on the HMM alone, but many of the action and especially object labels
attached to those clips are erroneous. Meanwhile the HMM-based procedure produces more reliable labels, but produces
fewer clips per video and suffers from occasional imprecision in determining the exact video start time to demonstrate an action's execution.

}

We cannot use keyword spotting if the goal is to align instructional
text to videos. However, if our goal is just to create a labeled
corpus of video clips, keyword spotting is a reasonable approach. Unfortunately, we
noticed that the quality of the labels (especially the object labels)
generated by keyword spotting was not very high, due to errors in the
ASR. On the other hand, we also noticed that the recall of the HMM approach
was about 5 times lower than using keyword spotting, and furthermore, that
the temporal localization accuracy was sometimes worse.

To get the best of both worlds, we employ the following hybrid technique.
We perform keyword spotting for the action in the ASR transcript as before,
but use the HMM alignment to infer the corresponding object.
\eat{
that is, we extract the object from the recipe step that is
aligned with the time window where the action occurs.
After performing this step, if there are any remaining
recipe steps that have not yet been associated to any video segments,
we use the standard HMM approach and extract a clip corresponding to
that step's segment, and derive its action and object label from the recipe text.
}
To avoid false positives, we only use the output of the HMM for this
video if at
least half of the recipe steps are aligned by it to the speech
transcript; otherwise we back off to the baseline approach of
extracting the noun phrase from the ASR transcript in the window after the verb.

\subsection{Temporal refinement using vision}
\label{sec:visualRefinement}
\label{sec:foodDetector}

\begin{figure*}[!t]
\centering
\includegraphics[width=0.95\textwidth]{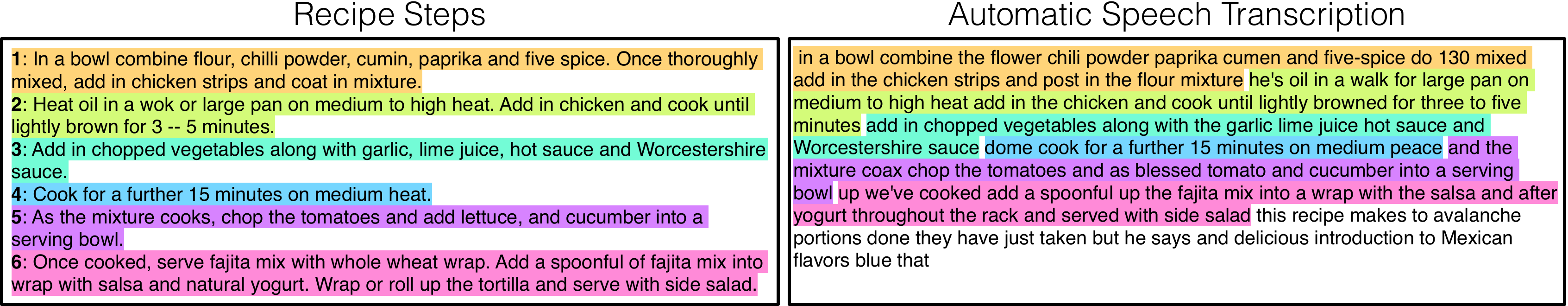}
\includegraphics[width=0.85\textwidth]{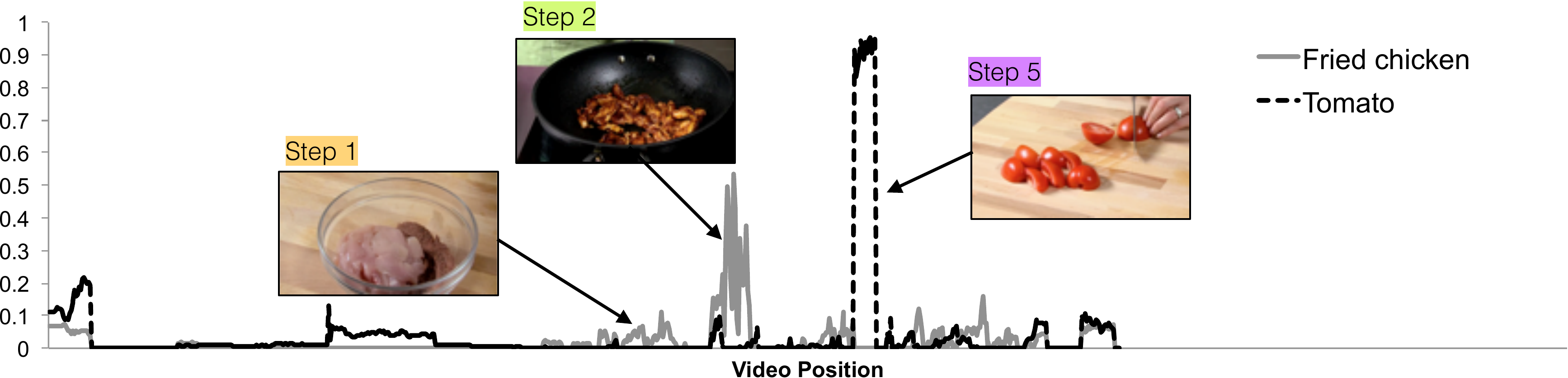}
\caption{\footnotesize
Examples from a Chicken Fajitas recipe at
\url{https://www.youtube.com/watch?v=mGpvZE3udQ4} (figure best
viewed in color).
Top: Alignment between (left) recipe
steps  to (right) automatic speech transcript.
Tokens from the ASR are allowed to be classified as background steps
(see e.g., the uncolored text at the end).
Bottom:
Detector scores for two ingredients
as a function of position in the video.
}
\label{fig:alignment_example_visual}
\label{fig:alignment_example}
\label{fig:foodDetectorTrace}
\end{figure*}

\eat{
\begin{figure*}[!t]
\centering
\subfigure[]{
\includegraphics[width=0.85\textwidth]{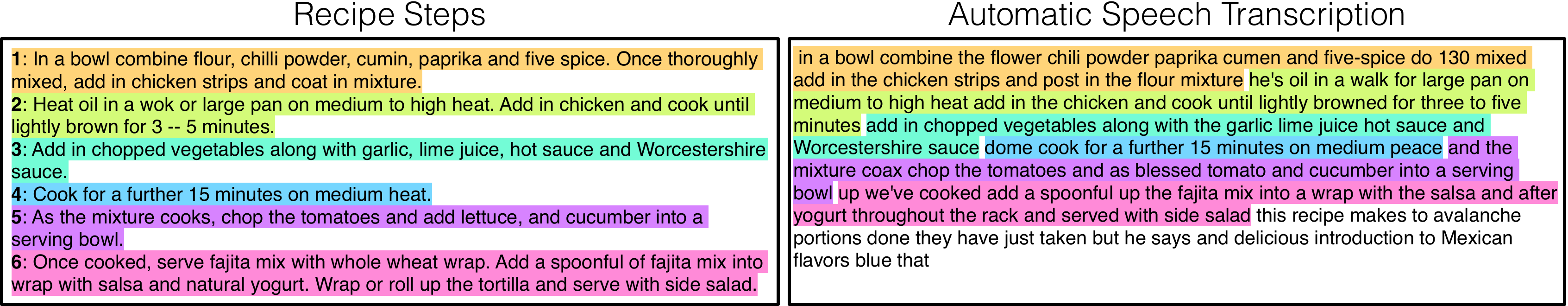}
\label{fig:alignment_example}
}
\subfigure[]{
\includegraphics[width=0.80\textwidth]{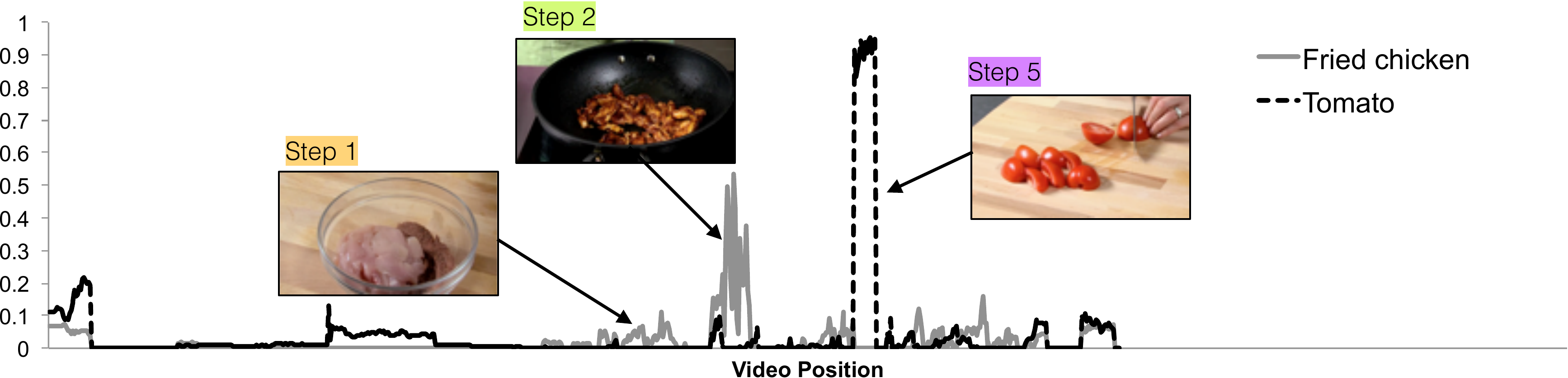}
\label{fig:alignment_example_visual}
}\vspace{-4mm}
\caption{\footnotesize
Examples from a Chicken Fajitas recipe at
\url{https://www.youtube.com/watch?v=mGpvZE3udQ4} (figures best
viewed in color);
\subref{fig:alignment_example}  Alignment between (left) recipe
steps  to (right) automatic speech transcript.
Tokens from the ASR are allowed to be classified as background steps; \subref{fig:alignment_example_visual}
Detector scores for two prominent recipe ingredients (of the same video)
as a function of position in
video with arrows depicting correspondence to recipe steps.\vspace{-3mm}
}
\label{fig:foodDetectorTrace}
\end{figure*}
}

In our experiments, we noticed that sometimes the narrator
describes an action before actually performing it (this was also noted in
\cite{Yu14}). To partially combat this problem, we used
computer vision to refine candidate video segments as follows.
We first trained visual detectors for a large collection of food items (described below).
Then, given a candidate video segment annotated with an action/object pair
(coming from any of the previous three
methods), we find a translation of the window (of up to 3 seconds in either direction)
for which the average detector score corresponding to the object is maximized.
The intuition is that by detecting when the
object in question is visually present in the scene, it is more likely
that the corresponding action is actually being performed.
\eat{
\footnote{
If we had reliable visual action detectors, we could use them as well.
Unfortunately,
 state of the art action recognition systems, such as \cite{Wang2013}
and those described in \cite{Thumos14}, require large labeled datasets
of examples for training. We are not aware of such datasets for the
domain of cooking, and indeed, one outcome of this paper is to create such a dataset.
}
}

\paragraph{Training visual food detectors.}

We trained a deep convolutional neural network (CNN) classifier
(specifically, the 16 layer VGG model from \cite{Simonyan2014})
on the FoodFood-101 dataset of~\cite{Bossard14},
using the Caffe open source software \cite{Jia2014}.
The Food-101 dataset contains 1000 images for 101 different kinds of
food.
To compensate for the small training set,
we pretrained the CNN on the ImageNet dataset
\cite{ILSVRCarxiv14}, which has 1.2M images, and then fine-tuned on Food-101.
After a few hours of fine tuning (using a single GPU),
we obtained 79\% classification
accuracy (assuming all 101 labels are mutually exclusive) on the test
set, which is consistent with the state of the art results.\footnote{
In particular, the website
\url{https://www.metamind.io/vision/food}
(accessed on 2/25/15)
claims they also got 79\% on
this dataset. This is much better than the 56.4\% for a CNN reported in
\cite{Bossard14}.
We believe the main reason for the improved performance is the use of
pre-training on ImageNet.
}

We then trained our model
on an internal, proprietary dataset of 220
million images harvested from Google Images and Flickr.
About 20\% of these images contain food, the rest are used to train the background class.
In this set, there are 2809 classes of food, including 1005 raw
ingredients, such as avocado or beef, and 1804 dishes, such as
ratatouille  or cheeseburger with bacon.
We use the model trained on this much larger dataset in the current
paper, due to its increased coverage.
(Unfortunately, we cannot report quantitative results,
since the dataset is very noisy (sometimes half
of the labels are wrong), so we have no ground truth.
Nevertheless, qualitative behavior is reasonable, and the model does
well on Food-101, as we discussed above.)

\paragraph{Visual refinement pipeline.}
For storage and time efficiency, we downsample each video temporally to 5 frames per second and
each frame to $224\times 224$ before applying the CNN.
 Running the food detector on each video then produces a
vector of scores (one entry for each of 2809 classes) per timeframe.

There is not a perfect map from the names of ingredients to the names of the
detector outputs.
For example, an omelette recipe may say ``egg'', but there are two kinds of
visual detectors, one for ``scrambled egg'' and one for ``raw egg''.
We therefore decided to define the match score
between an ingredient and a frame by taking the maximum score for that
frame over all detectors whose names matched any of the ingredient tokens
(after lemmatization and stopword filtering).

\eat{
Despite the large number of available categories,
associating mentioned objects from recipe text to detectors can be challenging
since some ingredient mentioned
might not correspond to any available detector, and even with an exact
string match, the mentioned object might be better associated with another detector.
For example, an omelette recipe might refer
to ``egg'' but we might be better off using a ``scrambled egg'' detector instead of a (raw) ``egg''
detector.  We thus match noun phrases to entire sets of foods which match the noun phrase
on any token after lemmatization and filtering for stopwords.
Our method of course has the problem of returning many false positives:
e.g., ``egg'' might match to ``scrambled egg'', ``egg custard'', ``egg wash'', and so on.
But we observe that so long as the false positives are not visually similar to other
ingredients in the same recipe, our accuracy is not negatively impacted.
}

Finally,
the match score of a video segment to an object is  computed
by taking the average score of all frames within that segment.
By then scoring and maximizing over all translations of the candidate
segment (of up to three seconds away), we produce a final ``refined'' segment.

\subsection{Quantifying confidence via vision and affordances}
\label{sec:confidence}

The output of the keyword spotting and/or HMM systems is an (action,
object) label assigned to certain video clips.
In order to estimate how much confidence we have in that label (so
that we can trade off precision and recall), we use
a linear combination of two quantities: (1) the final
match score produced by the visual refinement pipeline, which measures
the visibility of the object in the given video segment, and (2) an \emph{affordance
probability}, measuring the probability that $o$ appears as a direct object of $a$.

\begin{figure}[!t]
\centering
\includegraphics[width=.45\textwidth]{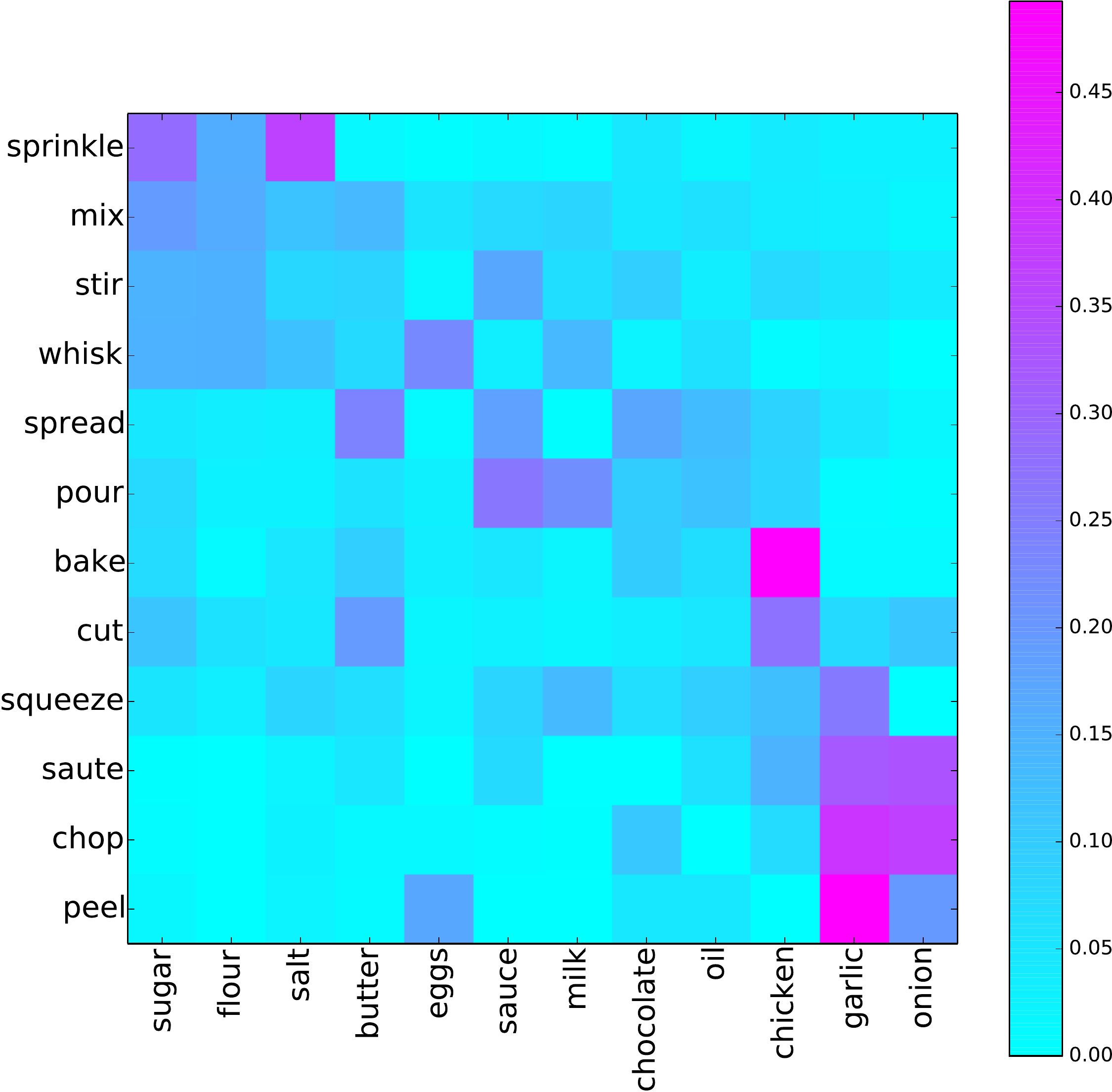}\vspace{-4mm}
\caption{\footnotesize Visualization of affordance model.  Entries $(a,o)$ are colored
according to $P(\mbox{object}=o\,|\,\mbox{action}=a)$.}\vspace{-2mm}
\label{fig:heatmap}
\end{figure}

The affordance model allows us to, for example, prioritize a segment
labeled as (peel, garlic) over a segment labeled as (peel, sugar).
The probabilities $P(\mbox{object}=o | \mbox{action}=a)$ are estimated
by first forming an inverse document frequency matrix capturing action/object
co-occurrences (treating actions as documents).  To generalize across actions
and objects we form a low-rank
approximation to this IDF matrix using a singular
value decomposition and set affordance probabilities to be
proportional to exponentiated entries
of the resulting matrix.
Figure~\ref{fig:heatmap} visualizes these  affordance probabilities for a selected
subset of frequently used action/object pairs.

\section{Evaluation and applications}
\label{sec:results}

In this section, we experimentally evaluate how well our methods work.
We then briefly demonstrate some prototype applications.

\eat{
We consider two main kinds of evaluation: ``macro reading'', in which
we evaluate the quality of the corpus of labeled clips that we have
created; and
``micro reading'', in which we evaluate the performance of individual
recipe/video alignments, using two different illustrative downstream applications.
}

\subsection{Evaluating the clip database}
\label{sec:clipEval}

\eat{
\begin{figure}[!t]
\centering
\includegraphics[width=0.4\textwidth,height=1.5in]{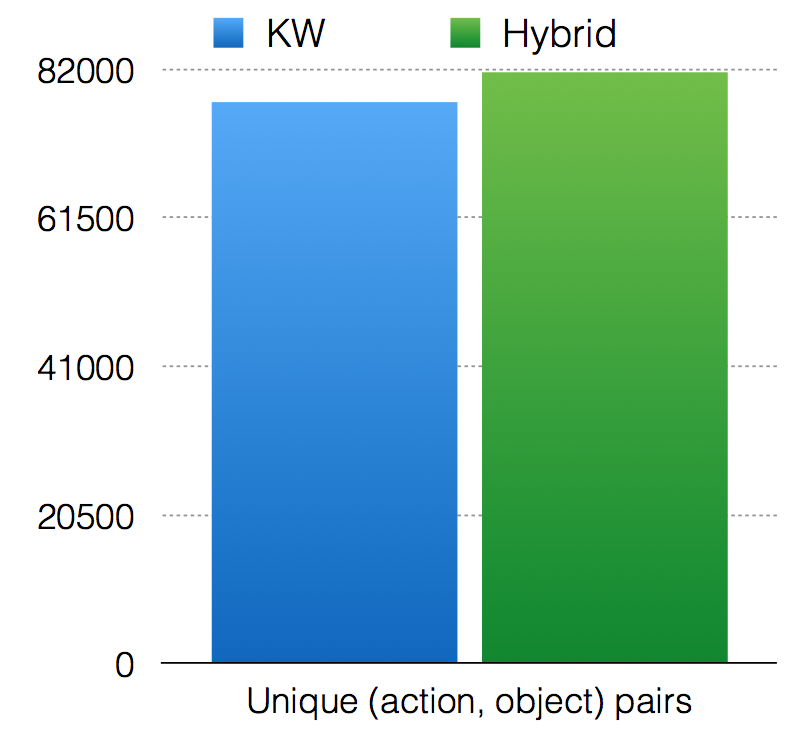}
\caption{Clip quantity}
\label{fig:clipQuantity}
\end{figure}
}

One of the main outcomes of our process is a set of video clips, each
of which is labeled with a verb (action) and a noun (object).
We generated 3 such labeled corpora, using 3 different methods:
keyword spotting (``KW''),
the hybrid HMM + keyword spotting (``Hybrid''),
and the hybrid system with visual food detector (``visual refinement'').
The total number of clips produced by each method is very similar,
approximately 1.4 million.
The coverage of the clips is approximately 260k unique (action, noun phrase) pairs.


\begin{figure}[!t]
\centering
\includegraphics[width=0.45\textwidth]{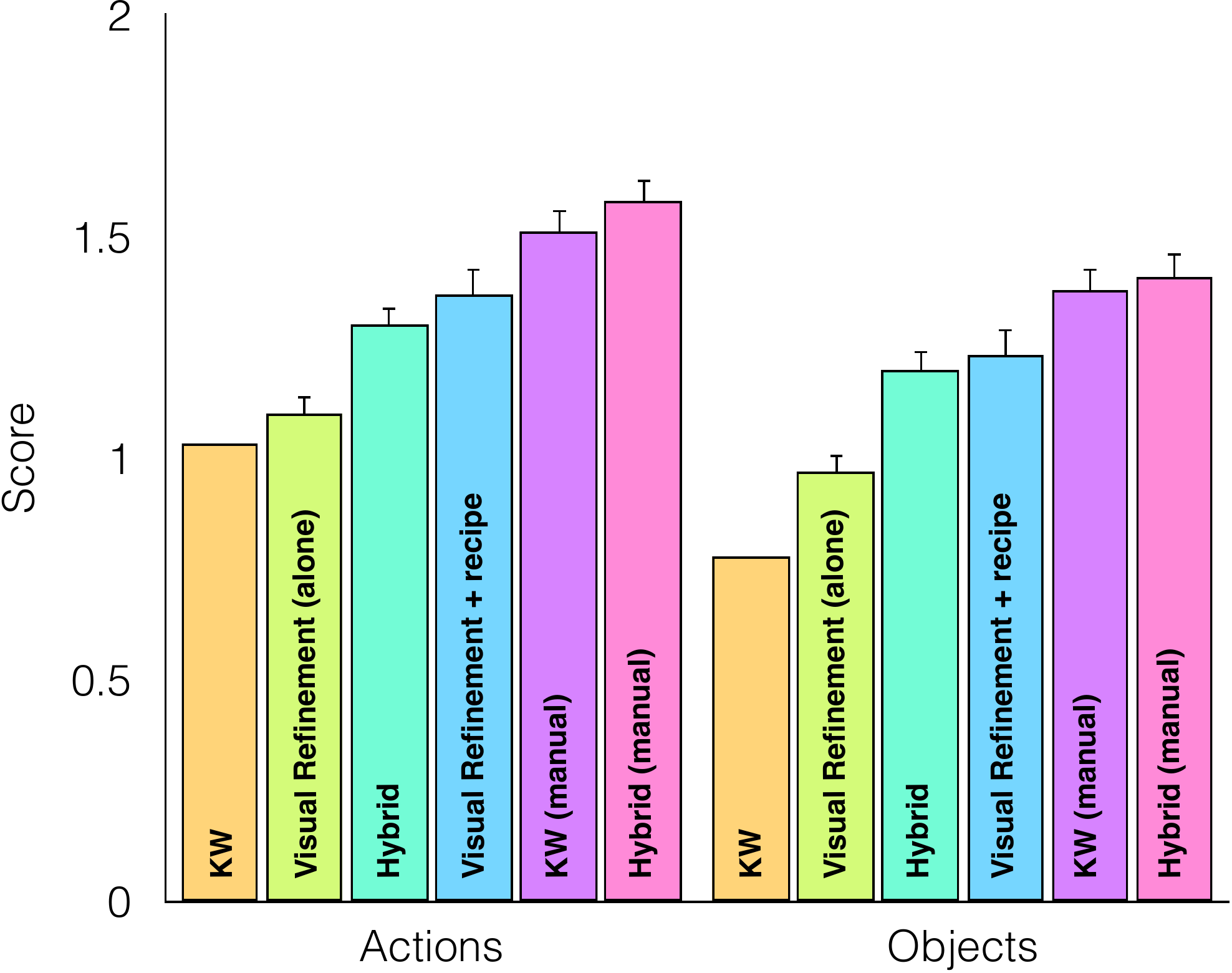}
\caption{\footnotesize
Clip quality, as assessed by Mechanical Turk experiments on
  900 trials.
Figure best viewed in color; see text for details.
}
\label{fig:clip_quality_barplot}
\end{figure}

\begin{figure}[!t]
\centering
\includegraphics[width=0.45\textwidth]{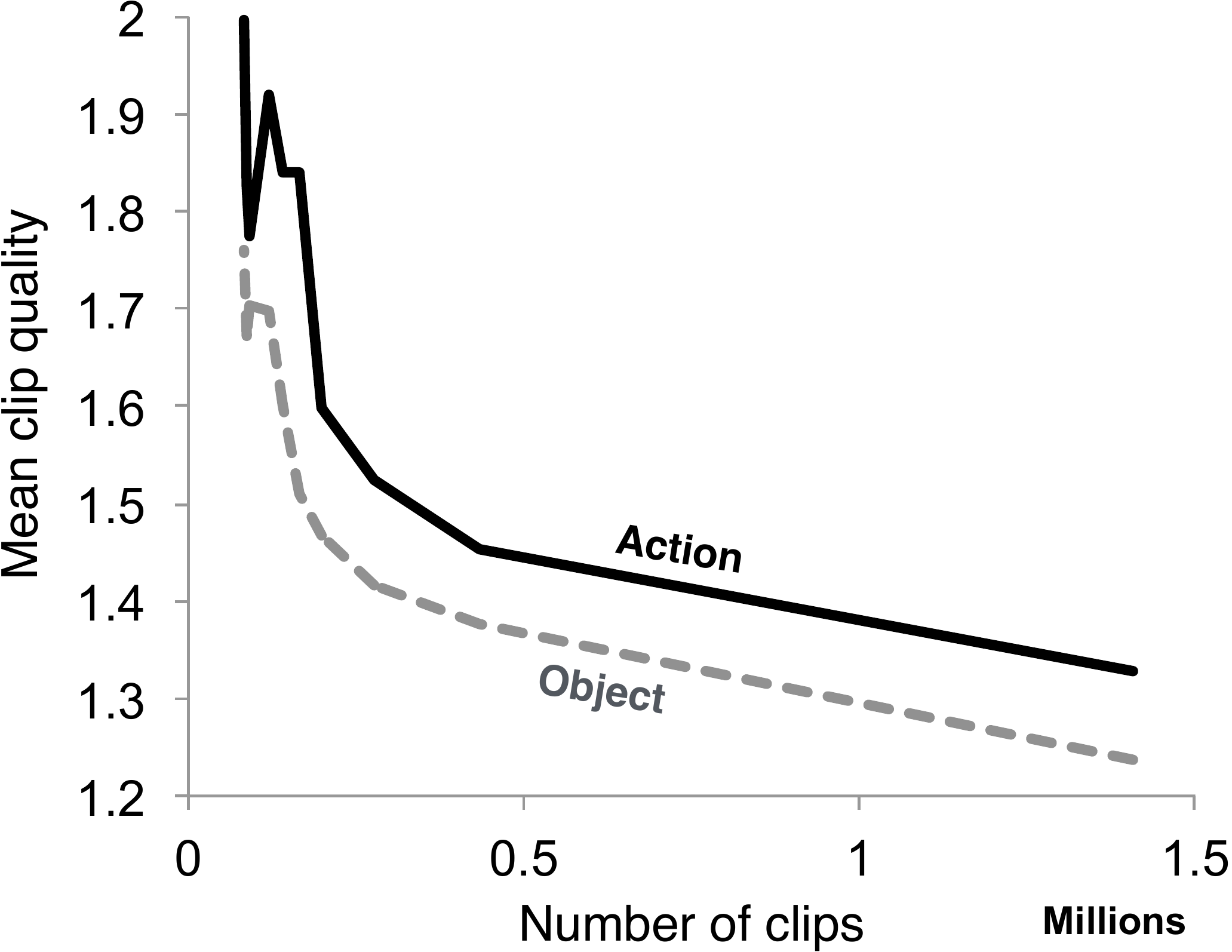}
\caption{\footnotesize
Average clip quality (precision) after filtering out low confidence clips
versus \# clips retained (recall).
}
\label{fig:confidence_roc}
\end{figure}

\begin{figure}[!t]
\centering
\includegraphics[width=0.3\textwidth]{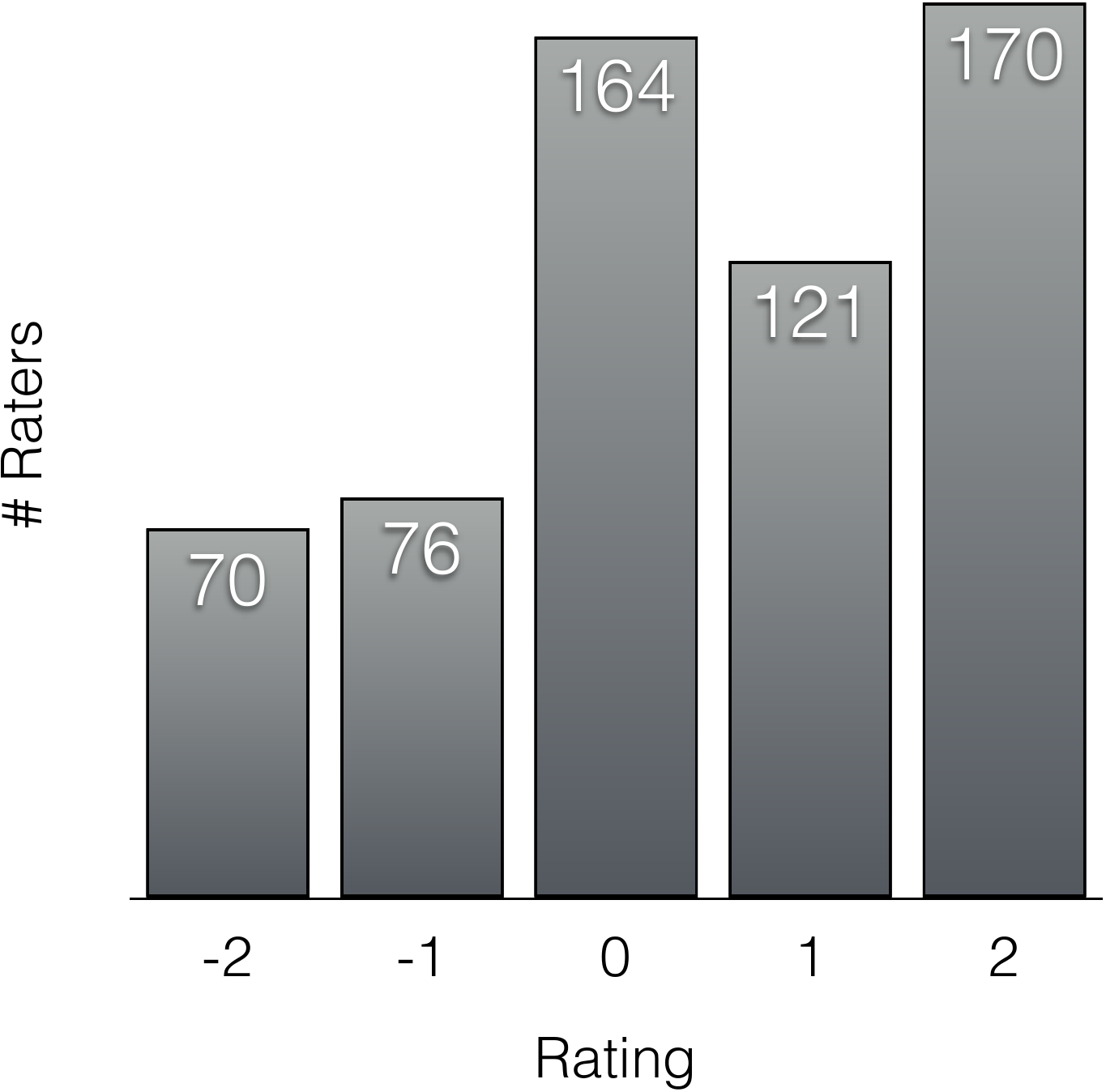}
\caption{\footnotesize
Histogram of human ratings comparing
recipe steps against ASR descriptions of a video clip. ``2'' indicate a strong preference
for the recipe step; ``-2'' a strong preference for the transcript. See text for details.
}
\label{fig:recipestep_v_asr}
\end{figure}

To evaluate the quality of these methods,
we created a random subset of 900 clips from each corpus using
stratified sampling. That is, we picked an action uniformly at random, and
then picked a corresponding object for that action from its support set uniformly at
random, and finally picked a clip with that (action, object) label uniformly
at random from the clip corpuses produced in Section~\ref{sec:methods}; this ensures the test set is not dominated by frequent
actions or objects.

We then performed a Mechanical Turk
experiment on each test set.
Each clip was shown to 3 raters, and each rater was asked the question
``How well does this clip show the given action/object?''.
Raters then had to answer  on a 3-point scale: 0 means ``not at all'',
1 means ``somewhat'', and 2 means ``very well''.

The results are shown in Figure~\ref{fig:clip_quality_barplot}.
We see that the quality of the hybrid method is significantly better
than the baseline keyword spotting method, for both actions and
objects.\footnote{
Inter-rater agreement, measured via Fleiss's kappa by aggregating across all judgment
tasks, is $.41$, which is statistically
significant at a $p<.05$ level.
}
While a manually curated speech transcript indeed yields better results (see the bars labeled `manual'),
we observe that automatically generated transcripts
 allow us to perform almost as well, especially using our alignment model with visual refinement.

Comparing accuracy on actions against that on objects in the same figure, we see that
keyword spotting is far more accurate for actions than it is for
objects (by over 30\%).
This disparity is not surprising since keyword spotting searches
only for action keywords and relies on a rough heuristic to recover objects.
We also see that using alignment (which extracts the object from the ``clean'' recipe text)
and visual refinement (which is trained explicitly to detect ingredients)
both help to increase the relative accuracy of objects --- under the hybrid method, for example,
the accuracy for actions is only 8\% better than that of objects.

Note that clips from the HMM and hybrid methods varied in length between 2 and 10 seconds (mean 4.2 seconds), while clips from the keyword spotting method were always exactly 8 seconds. Thus clip length is potentially a confounding factor in the evaluation when comparing the hybrid method to the keyword-spotting method; however, if there is a bias to assign higher ratings to longer clips (which are \textit{a priori} more likely to contain a depiction of a given action than shorter clips), it would benefit the keyword spoting method.

Segment confidence scores (from Section~\ref{sec:confidence}) can be used
to filter out low confidence segments, thus improving the precision of clip retrieval
at the cost of recall.  Figure~\ref{fig:confidence_roc} visualizes this trade-off as we
vary our confidence threshold, showing that indeed, segments with higher confidences
tend to have the highest quality as judged by our human raters.  Moreover, the top
167,000 segments as ranked by our confidence measure
have an average rating exceeding 1.75.

We additionally sought to evaluate how well recipe steps from the recipe body could serve as captions
for video clips in comparison to the often noisy ASR transript, which serves as a rough proxy
for evaluating the quality of the alignment model as well as demonstration a potential application of our method for ``cleaning up'' noisy ASR captions into complete grammatical sentences. To that end, we
randomly selected 200 clips from our corpus that both have an associated action keyword from the
transcript as well as an aligned recipe step selected by the HMM alignment model.
For each clip, three raters on Mechanical
Turk were shown the clip, the text from the recipe step, and a
fragment of the ASR transcript (the keyword, plus 5 tokens to the left
and right of the keyword). Raters then indicated which description
they preferred: 2 indicates a strong preference for the recipe step, 1
a weak preference, 0 indifference, -1 a weak preference for the
transcript fragment, and -2 a strong preference. Results are shown in
Figure~\ref{fig:recipestep_v_asr}. Excluding raters who indicated
indiffierence, 67\% of raters preferred the recipe step as the clip's description.

A potential confound for using this analysis as a proxy for the quality of the alignment model is that the ASR transcript is generally an ungrammatical sentence fragment as opposed to the grammatical recipe steps, which is likely to  reduce the raters' approval of ASR captions in the case when both accurately describe the scene. However, if users still on average prefer an ASR sentence fragment which describes the clip correctly versus a full recipe step which is unrelated to the scene, then this experiment still provides evidence of the quality of the alignment model.


\eat{
\subsection{Inferring object accordances}
\label{sec:affordances}

\begin{figure}[!t]
\centering
\includegraphics[height=2in,width=0.49\textwidth]{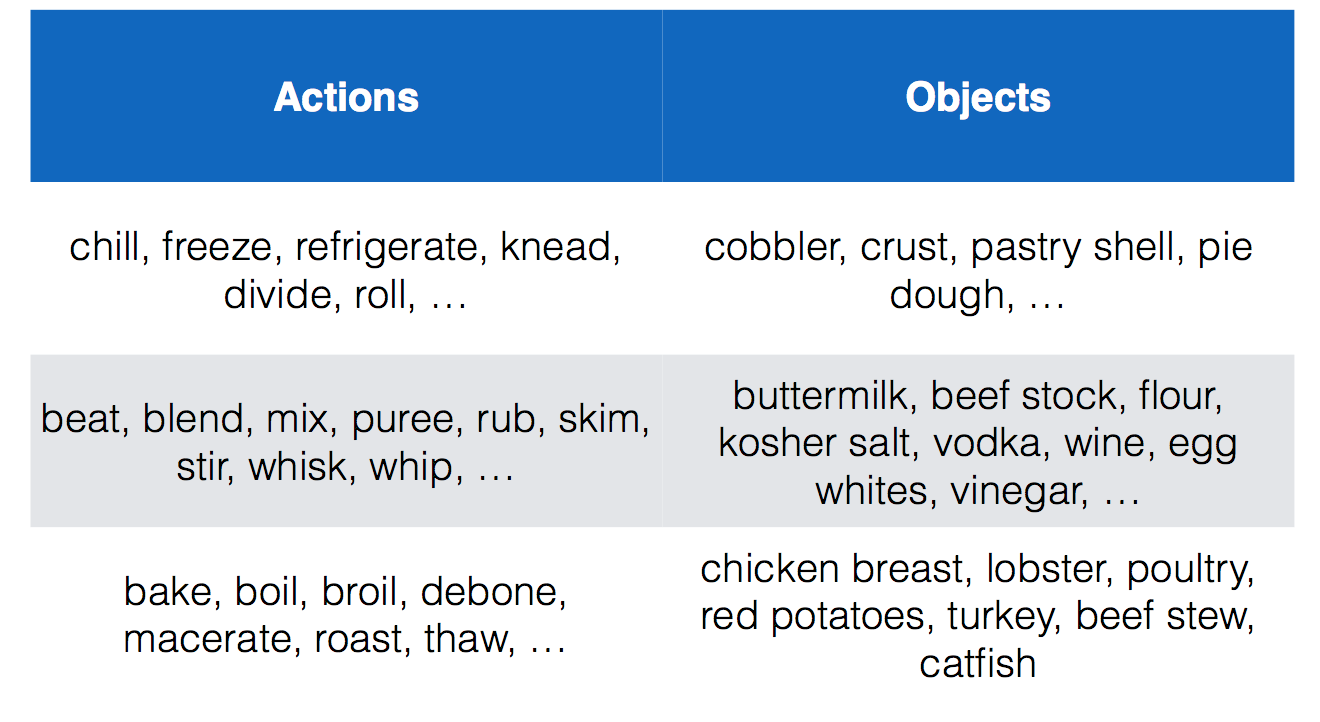}
\caption{Some of the biclusters derived from the action-object count matrix. See text for details.}
\label{fig:biclustering}
\end{figure}
}

\subsection{Automatically illustrating a recipe}
\label{sec:autoIllustrate}

\begin{figure*}[!t]
\centering
\includegraphics[width=.99\textwidth]{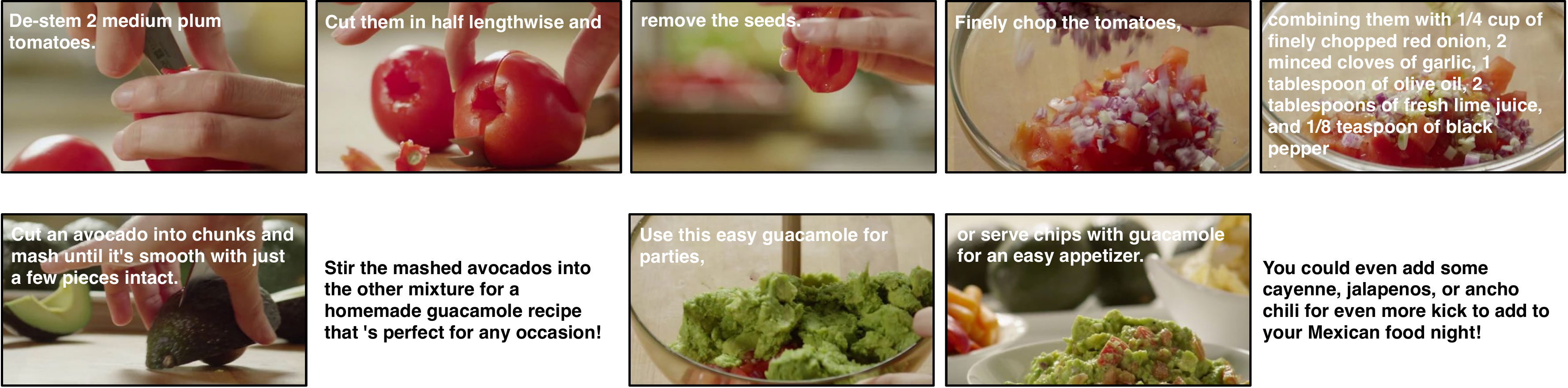}\vspace{-3mm}
\caption{\footnotesize Automatically illustrating a Guacamole recipe from \url{https://www.youtube.com/watch?v=H7Ne3s202lU}.}
\label{fig:autoIllustrateTomato}
\end{figure*}

One useful byproduct of our alignment method is that each recipe step
is  associated with a segment of the corresponding video.\footnote{
The HMM may assign multiple non-consecutive regions of the video to
the same recipe step (since the background state can turn on and off).
In such cases, we just take the ``convex hull'' of the regions as the interval
which corresponds to that step.
It is also possible for the HMM not to assign a given step to any
interval of the video.
} %
We use a standard keyframe selection algorithm to pick the best
frame from each segment. We can then associate this frame with the
corresponding recipe step, thus automatically illustrating the recipe steps.
An illustration of this process is shown in Figure~\ref{fig:autoIllustrateTomato}.

\subsection{Search within a video}
\label{sec:deepSearch}

\begin{figure}[!t]
\centering
\includegraphics[width=0.5\textwidth]{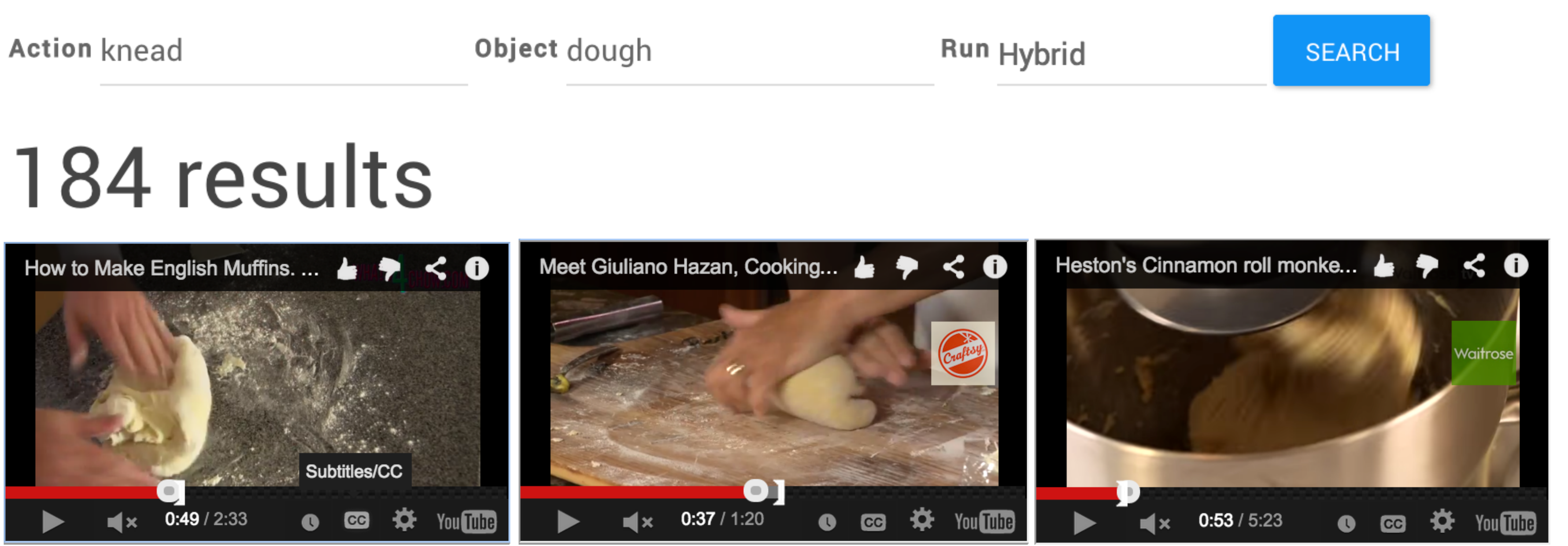}\vspace{-3mm}
\caption{\footnotesize Searching for ``knead dough''. Note that the videos have
  automatically been advanced to the relevant frame.
}\vspace{-3mm}
\label{fig:deepSearchRecipe}
\end{figure}

Another application which our methods enable is search within a
video. For example, if a user would like to find a clip illustrating how to
knead dough, we can simply search
our corpus of labeled clips, and return a list of matches
(ranked by confidence).  Since each clip has a corresponding ``provenance'', we can
return the results to the user as a set of videos in which we have
automatically ``fast forwarded'' to the relevant section of the video
(see Figure~\ref{fig:deepSearchRecipe} for an
example). This stands in contrast to standard video search on Youtube, which returns
the whole video, but does not (in general) indicate where within the video the user's
search query occurs.

\section{Related work}
\label{sec:related}

There are several pieces of related work.
\cite{Yu14} performs keyword spotting in the speech transcript
in order to label clips extracted from instructional videos.
However, our hybrid approach performs better; the gain is especially significant on automatically
generated speech transcripts, as shown in
Figure~\ref{fig:clip_quality_barplot}.

The idea of using an HMM to align instructional steps to a video
was also explored in \cite{Naim2014}.
However, their conditional model has to generate images,
whereas ours just has to generate ASR words, which is an easier task.
Furthermore, they only consider 6 videos collected in a controlled lab setting,
whereas we consider over 180k videos collected ``in the wild''.

Another paper that uses HMMs to process recipe text is \cite{Druck2012}.
They use the HMM to align the steps of a recipe to the comments made
by users in an online forum, whereas we align the steps of a recipe to
the speech transcript. Also, we use video information, which was not considered in this earlier work.

\cite{Joshi2006} describes a system to automatically illustrate a text
document, however they only generate one image, not a sequence, and
their techniques are very different.

\eat{
\cite{Bojanowski2014} describes a method to learn classifers to
perform action recognition, given a sequence of training labels but where the
precise temporal alignment to the video is unknown. They propose an
HMM-like model to perform the alignment. However, they do not use any
speech or text information, and the scale of their experiments is much
smaller than ours.
}

There is also a large body of other work on connecting language and
vision; we only have space to briefly mention a few key papers.
\cite{Rohrbach2012eccv} describes the MPII Cooking Composite Activities
dataset, which consists of 212 videos collected in the lab of people
performing various cooking activities. (This extends the dataset
described in their earlier work, \cite{Rohrbach2012cvpr}.)
They also describe a method to recognize objects and actions using
standard vision features.
However, they do not leverage the speech signal, and their dataset is significantly
smaller than ours.

\cite{Guadarrama2013} describes a method for generating
subject-verb-object triples given a short video clip,
using standard object and action detectors.
The technique was extended in
\cite{Thomason2014} to also predict the location/ place.
Furthermore, they use a linear-chain CRF to combine the visual
scores with a simple (s,v,o,p) language model (similar to our
affordance model).
They applied their technique to the dataset in \cite{Chen2011},
which consists of 2000 short video clips, each described with 1-3
sentences.
By contrast,  we focus on aligning instructional text to the video, and
our corpus is significantly larger.

\cite{Yu2013} describes a technique for estimating the compatibility
between a video clip and a sentence, based on relative motion of the
objects (which are tracked using HMMs).
Their method is tested on 159 video clips, created under carefully
controlled conditions.
By contrast,  we focus on aligning instructional text to the video, and
our corpus is significantly larger.

\section{Discussion and future work}
\label{sec:discuss}

In this paper, we have presented a novel method for aligning instructional text
to videos, leveraging both speech recognition and visual object
detection. We have used this to align 180k recipe-video
pairs, from which we have extracted a corpus of 1.4M labeled video clips -- a
small but crucial step toward building a multimodal procedural knowlege base.
In the future, we hope to use this labeled corpus to train visual
action detectors, which can then be combined with the existing visual
object detectors to interpret novel videos. Additionally, we believe that combining visual and linguistic cues may help overcome longstanding challenges to language understanding, such as anaphora resolution and word sense disambiguation.

\paragraph{Acknowledgments.} We would like to thank Alex Gorban and Anoop Korattikara for helping
with some of the experiments, and Nancy Chang for feedback on the paper.

\bibliographystyle{apalike}
\bibliography{refs}

\end{document}